\documentclass[conference]{IEEEtran}
\IEEEoverridecommandlockouts
\usepackage{cite}
\usepackage{amsmath,amssymb,amsfonts}
\usepackage{algorithmic}
\usepackage{graphicx}
\usepackage{textcomp}
\usepackage{xcolor}
\usepackage{subfigure}
\usepackage{ntheorem}
\def\BibTeX{{\rm B\kern-.05em{\sc i\kern-.025em b}\kern-.08em
    T\kern-.1667em\lower.7ex\hbox{E}\kern-.125emX}}
\theoremseparator{.}
\theorembodyfont{\upshape}
\def\IEEEproofindentspace{0\parindent}
\begin{document}

\title{\huge Analysis of Memory Capacity for Deep Echo State Networks}
\author{
\IEEEauthorblockN{Xuanlin Liu\IEEEauthorrefmark{1}, Mingzhe Chen\IEEEauthorrefmark{1}, Changchuan Yin\IEEEauthorrefmark{1} and Walid Saad\IEEEauthorrefmark{2}}
\IEEEauthorblockA{
\textit{\IEEEauthorrefmark{1}Beijing Key Laboratory of Network System Architecture and Convergence}\\
\textit{Beijing University of Posts and Telecommunications}, Beijing, China\\
Email: \{xuanlin.liu, chenmingzhe, ccyin\}@bupt.edu.cn.}
\IEEEauthorblockA{
\textit{\IEEEauthorrefmark{2}Wireless@VT, Bradley Department of Electrical and Computer Engineering}, 
\textit{Virginia Tech}, Blacksburg, VA, USA\\
Email: walids@vt.edu.}\vspace*{-1.5em}
\thanks{This research was supported by the U.S. National Science Foundation under Grants IIS-1838021 and CNS-1460316, and by the National Natural Science Foundation of China under Grant 61671086 and 61629101.}
}
\maketitle

\begin{abstract}
In this paper, the echo state network (ESN) memory capacity, which represents the amount of input data an ESN can store, is analyzed for a new type of deep ESNs.
In particular, two deep ESN architectures are studied. First, a parallel deep ESN is proposed in which multiple reservoirs are connected in parallel allowing them to average outputs of multiple ESNs, thus decreasing the prediction error. Then, a series architecture ESN is proposed in which ESN reservoirs are placed in cascade that the output of each ESN is the input of the next ESN in the series. This series ESN architecture can capture more features between the input sequence and the output sequence thus improving the overall prediction accuracy. Fundamental analysis shows that the memory capacity of parallel ESNs is equivalent to that of a traditional shallow ESN, while the memory capacity of series ESNs is smaller than that of a traditional shallow ESN.
%As the size of the data that ESN can record increases, the ESN can approximate more complex systems accurately. The deep ESNs in parallel architecture and in series architecture are proposed to overcome the traditional ESN disadvantages of possible huge prediction deviation and low complex system prediction accuracy. The memory capacity of deep ESNs is analyzed, which shows that it depends on reservoir size and the reservoir weight.% and the complexity of the system being predicted. 
In terms of normalized root mean square error, simulation results show that the parallel deep ESN achieves 38.5\% reduction compared to the traditional shallow ESN while the series deep ESN achieves 16.8\% reduction.
\end{abstract}

\section{Introduction}
Reservoir computing (RC) is a state-space model that follows a fixed state transition structure (known as a \textit{reservoir}) and an adaptable output \cite{Rodan2011Minimum}. Echo state network (ESN) is considered as one of the simplest forms of the RC model that is useful for processing time series. An ESN typically uses ten times less neurons compared to other recurrent neural networks (RNNs).
Furthermore, only the output weight matrix in an ESN needs to be trained \cite{Jaeger2004Harnessing}.
%and only the weight matrix between the reservoir and the output are modified during training, instead of all connections in the RNNs \cite{Jaeger2004Harnessing}. 
Due to their effectiveness and training simplicity, ESNs have been widely used in many fields \cite{chen2017machine,Abadlia2017ESN,Chen2017Caching,Pugach2017UAV,7412759,Pan2017traffic,Peng2017ship,Jaeger2004Harnessing,Chen2018TC}, including time series prediction, wireless networks, and unmanned aerial vehicle control and communication. However, due to the simple structure and randomness of the weight matrices, ESN faces many challenges that include minimizing prediction errors and enhancing prediction accuracy for highly complex systems.
%prediction error and the incompatibility with the complexity of the predicted system. The incompatibility indicates that the prediction accuracy of an ESN decreases rapidly as the system becomes more complex.

The existing literature such as \cite{Rodan2010Simple,Rodan2012Simple,Peng2017ship,Pan2017traffic,Deng2006Complex,Deng2016Deep,Jaeger2001Short} has studied a number of problems related to ESNs. The authors in \cite{Pan2017traffic} introduced a bat algorithm to overcome the influences of initial random weights, thereby improving the effectiveness and robustness of the ESN prediction system. The work in \cite{Peng2017ship} proposed a Kalman filter to improve ESN predictions by recursively training the network output weight.
%The authors designed simple cycle reservoir in \cite{Rodan2010Simple} and cycle reservoir with jumps in \cite{Rodan2012Simple}. 
The authors in \cite{Rodan2010Simple} and \cite{Rodan2012Simple} proposed simple cycle reservoirs and cycle reservoirs with jumps so as to shorten the trail session for a specific system. The authors in \cite{Deng2006Complex} proposed a scale-free highly-clustered ESN (SNESN) to capture a large number of features of the ESN input stream for better predictions. In \cite{Deng2016Deep}, the authors proposed a deep self-organizing SHESN so as to construct a large system with a stack of well-trained reservoirs, which improves the prediction ability of the network. However, most of the existing literature such as \cite{Rodan2010Simple,Rodan2012Simple,Peng2017ship,Pan2017traffic,Deng2006Complex,Deng2016Deep} only tested the prediction capability of ESNs with various datasets but did not focus on the theoretical analysis. The author in \cite{Jaeger2001Short} introduced the concept of a short-term memory capacity of ESN to provide a quantitative measure of the prediction capability. The work in \cite{Chen2017ESN} proposed an ESN-based algorithm and analyzed the memory capacity of the ESN to predict the content request distribution and mobility pattern for mobile users in cloud radio access networks. However, the work in \cite{Jaeger2001Short} and \cite{Chen2017ESN} did not propose a structural solution to minimizing prediction errors stemming from ESN, particularly when dealing with highly complex systems.

The main contribution of this paper is to propose new architectures of ESNs and evaluate their capability of recording the historical data. \textit{To our best knowledge, this is the first work that analyzes the prediction capability of deep ESNs.} In this regard, our key contributions include:
\vspace{-0.1em}
\begin{itemize}
    \item To improve and stabilize the prediction accuracy of ESN, we propose deep ESN architectures composed of multiple reservoirs in parallel-connection and series-connection, respectively. The parallel ESN decreases the prediction error by averaging outputs of multiple ESNs. The series ESN captures more features for complex system than the traditional shallow ESN and improves the prediction accuracy.
    \item We analyze the memory capacity \cite{Jaeger2001Short} of deep ESNs and provide a measure to evaluate the historical data memory of deep ESNs. The parallel architecture keeps the memory capacity of the traditional shallow ESN for recording historical data, while the series architecture misses more historical data than the shallow ESN.
    %while the series architecture introduces a penalty for expanding the network in cascading. The relevance between input streams is also a factor of memory capacity.
    \item Simulation results show that the normalized root mean square error (NRMSE) is reduced by 38.5\% in the parallel ESN and 16.8\% in the series ESN compared with the traditional shallow ESN.
\end{itemize}

The rest of this paper is organized as follows. The preliminaries of ESN and the proposed parallel and series deep ESN architectures are introduced in Section~\ref{sec:enESN}. In Section~\ref{sec:MC}, the memory capacity of deep ESNs is analyzed. Numerical simulation results are presented and analyzed in Section~\ref{sec:exp}. Finally, conclusions are drawn in Section~\ref{sec:con}.

\section{Deep ESNs}\label{sec:enESN}
In this section, we first introduce the architecture of a traditional shallow ESN. To improve the prediction accuracy of ESNs, we propose two novel deep architectures that rely on ESN: a parallel architecture and a series architecture.

\subsection{Echo State Networks: Preliminaries}
\begin{figure}[tbp]
\centerline{\includegraphics[width=3in]{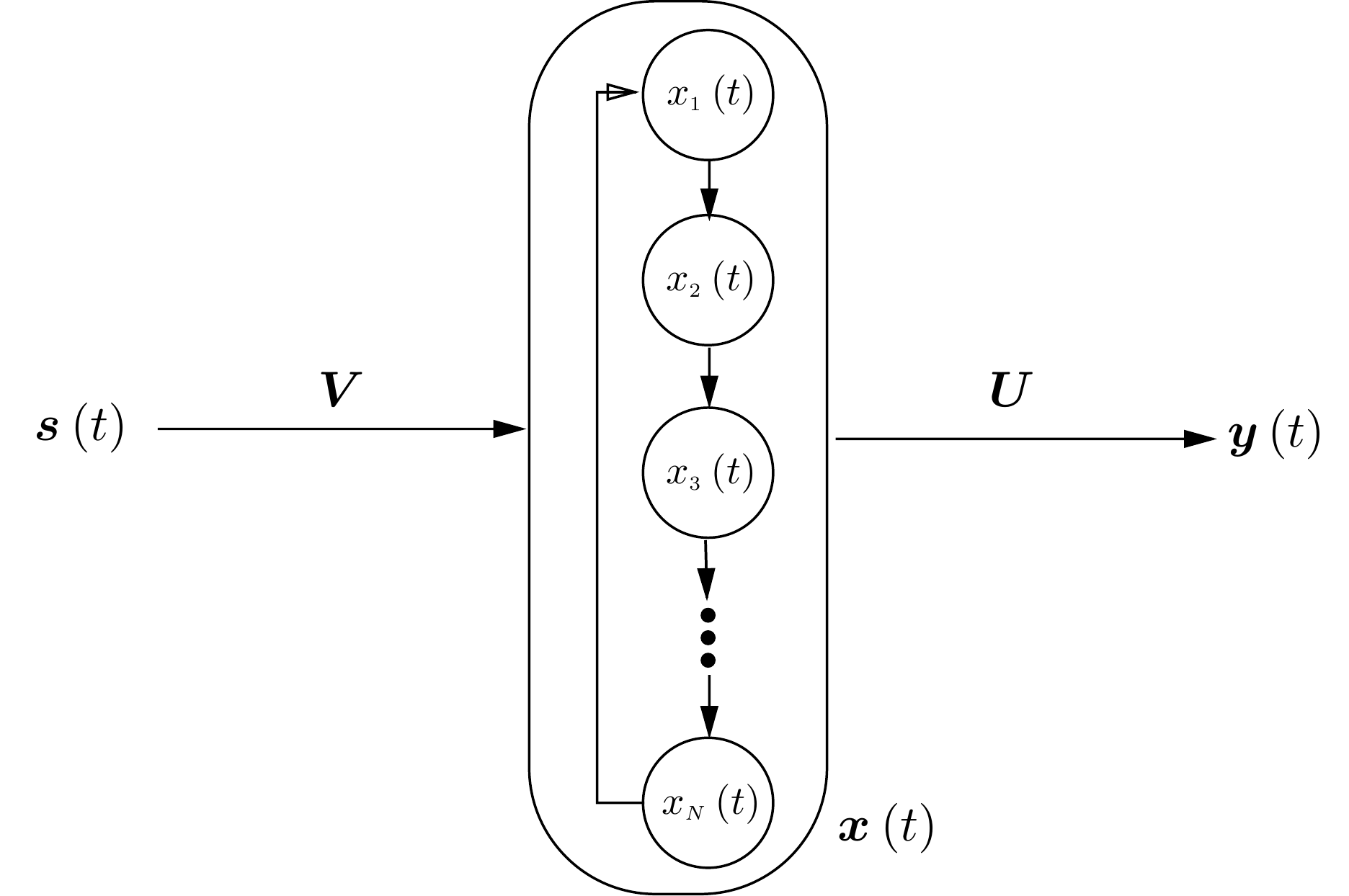}}
\caption{The echo state network architecture.}
\label{figure:ESN}
\end{figure}
An ESN consists of $K$ input units, $N$ reservoir units, and $M$ output units, as shown in \figurename~\ref{figure:ESN}. The activations of input, reservoir, and output units at time $t$ are given by $\boldsymbol s\left(t\right)$, $\boldsymbol x\left(t\right)$, and $\boldsymbol y\left(t\right)$, respectively. The input matrix $\boldsymbol V\in \mathbb R ^{N\times K}$ represents the transformation from the input units to the reservoir units. The reservoir updating matrix $\boldsymbol W\in \mathbb R ^{N\times N}$ represents the updating rule of the reservoir units over time. The output matrix $\boldsymbol U\in \mathbb R ^{M\times N}$ represents the transformation from reservoir to output units.

$\boldsymbol V$ and $\boldsymbol W$ are constant matrices whose elements take values that are generated randomly in $\left(-1,1\right)$ before the training of ESNs. The reservoir updating matrix $\boldsymbol W$ is scaled as follows: $\boldsymbol W\leftarrow \alpha \boldsymbol W/\left| \lambda _{\textrm{max}}\right|$, where $\left| \lambda _{\textrm{max}}\right|$ is the spectral radius of $\boldsymbol W$ and $\alpha \in \left( 0,1\right)$ is a scaling parameter \cite{Jaeger2002Tutorial}. The reservoir units and output units are updated with time $t$ as follows:
\begin{equation}
    \boldsymbol x\left( t+1\right) =f\left(\boldsymbol V\boldsymbol s\left( t+1\right) +\boldsymbol W\boldsymbol x\left( t\right) \right)\label{reservoir_update}{\textrm{,}}
\end{equation}
\begin{equation}
    \boldsymbol y\left( t+1\right) =\boldsymbol U\boldsymbol x\left( t+1\right) \label{output}{\textrm{,}}
\end{equation}
where $f\left( \cdot \right)$ is the activation function, $f\left(\cdot\right)$ can be typically defined as a sigmoid or $tanh$ function \cite{Rodan2011Minimum}. 

An ESN can be trained offline and the output weight matrix $\boldsymbol U$ is calculated using ridge regression \cite{Wyffels2008Stable} as follows:
\begin{equation}
    \boldsymbol U=\left( \boldsymbol X^{\textrm{T}}\boldsymbol X+\lambda^2\boldsymbol I \right)^{-1}\boldsymbol X^{\textrm{T}}\boldsymbol y\label{ridge_regression}{\textrm{,}}
\end{equation}
where $\boldsymbol X$ holds the reservoir states, $\boldsymbol I$ is the identity matrix, $\lambda$ is the regularization factor greater than 0, and $\boldsymbol y$ is a vector of output of training sequence. Since ESN is one of the simplest forms of reservoir computing, the prediction accuracy is affected by the architecture of the ESN, which is generated randomly, as well as, the complexity of the predicted system. Next, we introduce the deep ESNs as a structural solution to improve the prediction accuracy.

\subsection{Parallel ESN}
\begin{figure}[tbp]
\centerline{\includegraphics[width=6.5cm]{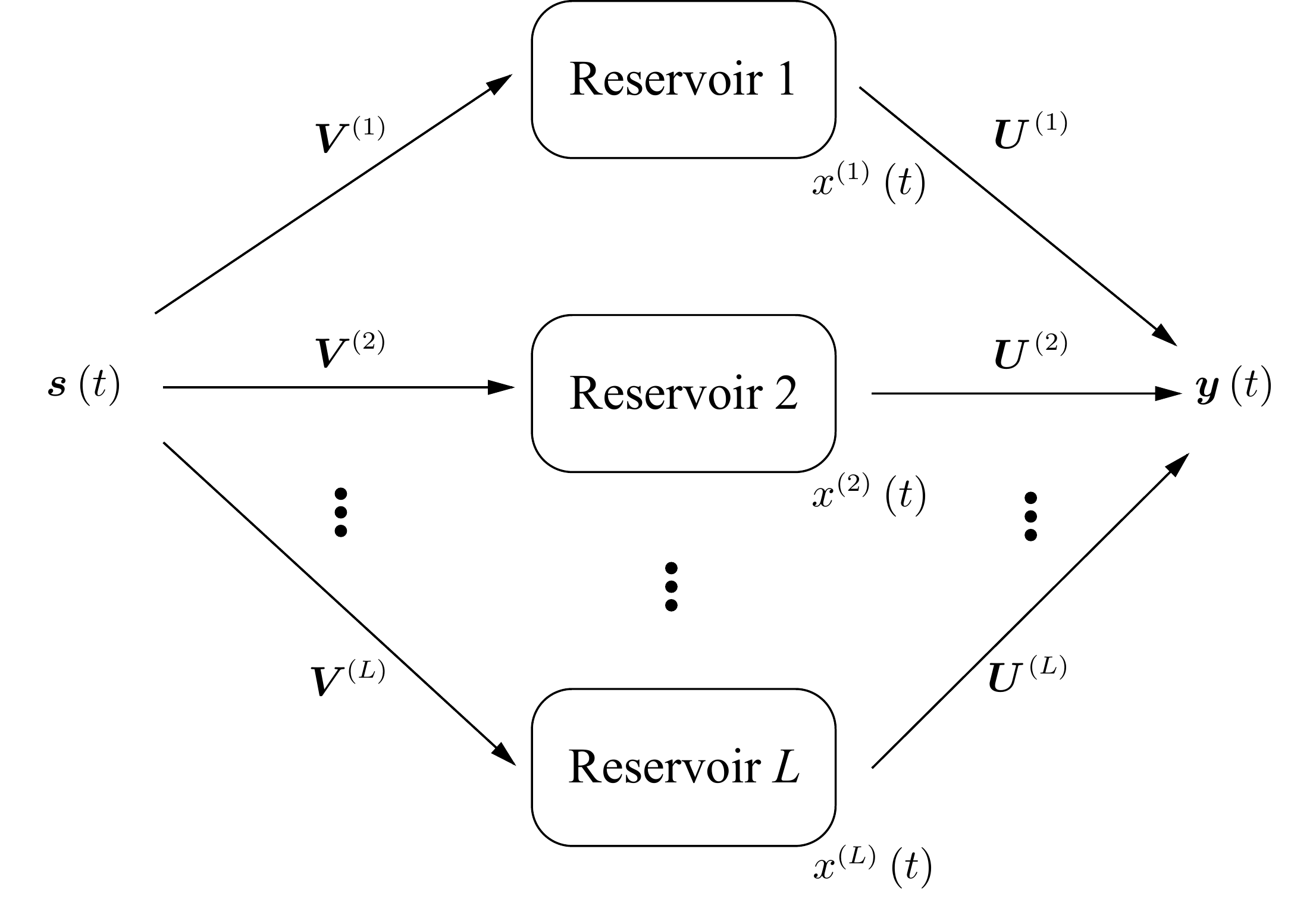}}
\caption{Parallel ESN architecture with $L$ reservoirs.}
\label{figure:parallel}
\end{figure}
The architecture of a parallel ESN is shown in \figurename~\ref{figure:parallel}. An input sequence $\boldsymbol s\left( t\right)$ enters $L$ reservoirs simultaneously. $\boldsymbol V^{\left( l\right)}$ and $\boldsymbol U^{\left( l\right)}$ represent the input matrix and output matrix for reservoir,~$l\in \{ 1,2,\ldots ,L\}$. The input matrices and reservoir updating matrices are considered to be constant and generated randomly before training. The training process is similar to that of shallow ESNs and all $L$ reservoirs can be trained simultaneously. The output matrices of $L$ reservoirs are determined after training. The units in $L$ reservoirs and the output units are updated as follows:
\begin{equation}
    \boldsymbol x^{\left(l\right)}\left( t+1\right) =f\left(\boldsymbol V^{\left(l\right)}\boldsymbol s\left( t+1\right) +\boldsymbol W^{\left(l\right)}\boldsymbol x^{\left(l\right)}\left( t\right) \right)\label{reservoir_update_parallel}{\textrm{,}}
\end{equation}
\begin{equation}
    \boldsymbol y^{\left(l\right)}\left( t+1\right) =\boldsymbol U^{\left(l\right)}\boldsymbol x^{\left(l\right)}\left( t+1\right) \label{output_parallel}{\textrm{.}}
\end{equation}

The output of a parallel ESN is the arithmetic mean of $L$ reservoir outputs, which is given by:
\begin{equation}
    \boldsymbol y\left(t\right)=\sum_{l=1}^{L}\boldsymbol y^{\left(l\right)}\left(t\right)/L{\textrm{.}}
\end{equation}

By taking the average of $L$ reservoir outputs, the parallel architecture decreases the prediction error and improves the accuracy. Furthermore, compared to a shallow ESN with $L\times N$ neurons, a parallel ESN with $L$ reservoirs costs less in training since it only needs to train $L$ output matrices of $M\times N$ instead of a relatively large output matrix of $M\times LN$.

\subsection{Series ESN}
\begin{figure*}[tbp]
\centerline{\includegraphics[width=6.2in]{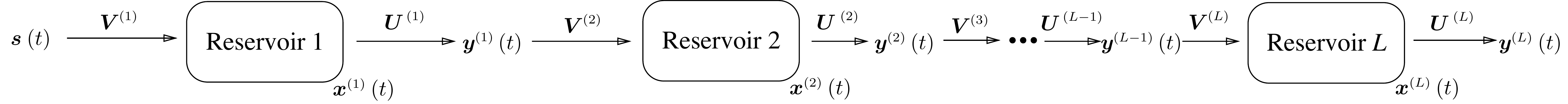}}
\caption{Series ESN architecture with $L$ reservoirs.}
\label{figure:series}
\end{figure*}
A series ESN consists of $L$ reservoirs, as shown in \figurename~\ref{figure:series}. An input sequence $\boldsymbol s\left( t\right)$ enters into the first reservoir. The output of the previous reservoir will be the input of the following one. Similarly, the input matrices $\boldsymbol V^{\left( l\right)}$ and reservoir updating matrices $\boldsymbol W^{\left( l\right)}$ are constant and generated randomly before training. A series ESN is trained sequentially. The first reservoir is trained to predict the output sequence based on the input sequence. $L$ reservoirs are all trained to predict the output of a nonlinear system. The difference between training a series ESN and a parallel ESN is that the first reservoir is trained with the input of the system, while any subsequent reservoir is trained with the output of the previous reservoir. The output matrices $\boldsymbol U^{\left( l\right)}$ are determined after training. Units in reservoir $l$ and the output of reservoir $l$ are given by:
\begin{equation}
    \boldsymbol x^{\left(l\right)}\left( t+1\right) =f\left(\boldsymbol V^{\left(l\right)}\boldsymbol y^{\left(l-1\right)}\left( t+1\right) +\boldsymbol W^{\left(l\right)}\boldsymbol x^{\left(l\right)}\left( t\right) \right)\label{reservoir_update-series}{\textrm{,}}
\end{equation}
\begin{equation}
    \boldsymbol y^{\left(l\right)}\left( t+1\right) =\boldsymbol U^{\left(l\right)}\boldsymbol x^{\left(l\right)}\left( t+1\right) \label{output-series}{\textrm{,}}
\end{equation}
where $l \in \{1,2,\ldots,L\}$ and $\boldsymbol y^{\left(0\right)}\left(t\right)=\boldsymbol s\left(t\right)$. $\boldsymbol y^{\left(L\right)}\left(t\right)$ is the output $\boldsymbol y\left(t\right)$ of a series ESN.

\section{Short-Term Memory Capacity of Deep ESN Architecture}\label{sec:MC}
In this section, we analyze the short-term memory capacity (MC) of deep ESNs. The short-term MC is used to quantify the memory capability of a recurrent network architecture for recording information from the input stream \cite{Rodan2011Minimum}. 

The MC is defined as the squared correlation coefficient between the desired output ($k$-time-step delayed input signal, $\boldsymbol s\left( t-k\right)$) and the observed network output $\boldsymbol y\left( t\right)$, which can be given as \cite{Jaeger2001Short}:
\begin{equation}
    C_k=\frac{\textrm{Cov}^2\left( \boldsymbol s\left( t-k\right),\boldsymbol y\left( t\right) \right)}{\textrm{Var}\left( \boldsymbol s\left( t\right)\right)\textrm{Var}\left(\boldsymbol y\left(t\right)\right)}\label{MC_k}{\textrm{,}}
\end{equation}
where $\textrm{Cov}\left(\cdot\right)$ and $\textrm{Var}\left(\cdot\right)$ denote the covariance and variance operators, respectively. The short-term MC is defined as:
\begin{equation}
    C = \sum_{k=1}^{\infty}C_k\label{MC}{\textrm{.}}
\end{equation}

To quantify the MC of deep ESNs, we first assume that the reservoir updating matrix is given by:
\begin{equation}\label{scr_w}
\boldsymbol W=\left[\begin{matrix}
0 & 0 & \cdots & r\\
r & 0 & 0 & 0\\
0 & \ddots & 0 & 0\\
0 & 0 & r & 0
\end{matrix}\right]_{N\times N}{\textrm{,}}
\end{equation}
where $r$ represents the reservoir weight, which is set to a constant value prior to ESN training. Based on \cite[Appendix B]{Rodan2011Minimum}, we define auxiliaries to simplify the derivation as follows: 1) the rotation operator $\textrm{rot}_k\left(\cdot\right)$ used for cyclically rotating the elements in a $k$ places to the right; 2) the extension matrix of $\boldsymbol V$, $\Omega=\left({\textrm{rot}}_1\left(\boldsymbol V_{N\ldots1}\right),{\textrm{rot}}_2\left(\boldsymbol V_{N\ldots1}\right),\ldots,{\textrm{rot}}_N\left(\boldsymbol V_{N\ldots1}\right)\right)$; 3) the diagonal matrix $\Gamma={\textrm{diag}}\left(1,r,r^2,\ldots,r^{N-1}\right)$; 4) the invertible matrix ${\boldsymbol A}=\Omega^{\textrm{T}}\Gamma^2\Omega$. Furthermore, we leverage the following result~\cite{Rodan2011Minimum}: $\zeta_k=\left(\textrm{rot}_k\left(\boldsymbol V_{1\dots N}\right)\right)^\textrm{T}\boldsymbol A^{-1}\textrm{rot}_k\left(\boldsymbol V_{1\dots N}\right)=r^{-2k}$, $k\in\left\{0,1,2,\ldots,N-1\right\}$. We can also deduce that $\left(\textrm{rot}_i\left(\boldsymbol V_{1\dots N}\right)\right)^\textrm{T}\boldsymbol A^{-1}\textrm{rot}_j\left(\boldsymbol V_{1\dots N}\right)=0$, $i,j\in\left\{0,1,2,\ldots,N-1\right\}$ and $i\neq j$. 
Then, the MC of a parallel ESN can be given by the following theorem.
\newtheorem{theorem}{\bf Theorem}
\begin{theorem}\label{theorem:parallel}
  In a parallel ESN which consists of $L$ single ESNs connected in parallel, we assume that each single ESN's input matrix $\boldsymbol V^{\left(l\right)}$ guarantees a regular matrix $\Omega^{\left(l\right)}$, $l\in \{ 1,2,\ldots ,L\}$. Then, the MC of the parallel ESN is $C = N-1+r^{2N}$.
  % \begin{equation}
  %   C = N-1+r^{2N}\label{parallel_MC}{\textrm{.}}
  % \end{equation}
\end{theorem}
\begin{IEEEproof}
    The input stream $s\left(t\right)$ is zero-mean real-valued. The activations of units in reservoir $l$ at time $t$ are given by:
    \begin{equation}
    \boldsymbol x^{\left(l\right)}\left(t\right)=\sum_{i=0}^{\infty}{{\textrm{rot}}_{i}\left(\boldsymbol V_{1\ldots N}^{\left(l\right)}\right)}r^is\left(t-i\right){\textrm{.}}\label{parallel_x}
    \end{equation}

    The output matrix of reservoir $l$ is ${\boldsymbol U^{\left(l\right)}}=\left({\boldsymbol R^{\left(l\right)}}\right)^{-1}\boldsymbol p_k^{\left(l\right)}$, where the covariance matrix of reservoir $l$'s activations is:
    \begin{equation}
    \begin{aligned}
     {\boldsymbol R}^{\left(l\right)}&={\mathbb E}\left[\boldsymbol x^{\left(l\right)}\left(t\right)\left({\boldsymbol x^{\left(l\right)}}\left(t\right)\right)^{\textrm{T}}\right]\\
     &=\frac{\sigma ^2}{1-r^{2N}}\left(\Omega^{\left(l\right)}\right)^{\textrm{T}}\Gamma ^2\Omega ^{\left(l\right)}=\frac{\sigma ^2}{1-r^{2N}}{\boldsymbol A}^{\left(l\right)}\label{parallel_R}{\textrm{,}}
    \end{aligned}
    \end{equation}
    and the expectation of the product of reservoir $l$'s activations and $k$-slot delayed source is given by:
  \begin{equation}
  \begin{aligned}
    \boldsymbol p_k^{\left(l\right)}={\mathbb E}\left[\boldsymbol x^{\left(l\right)}\left(t\right)s\left(t-k\right)\right]=\sigma ^2r^{k}{\textrm{rot}}_k\left(\boldsymbol V_{1\ldots N}^{\left(l\right)}\right)\label{parallel_pk}{\textrm{.}}
    \end{aligned}
  \end{equation}
  
  Hence, the output matrix of reservoir $l$ will be:
  \begin{equation}
    {\boldsymbol U}^{\left(l\right)}=\left(1-r^{2N}\right)r^{k}\left({\boldsymbol A}^{\left(l\right)}\right)^{\left(-1\right)}{\textrm{rot}}_{k}\left(\boldsymbol V_{1\ldots N}^{\left(l\right)}\right){\textrm{.}}
  \end{equation}
  
  The output at time $t$ is given by:
  \begin{displaymath}
    \begin{aligned}
        y^{\left(l\right)}\left(t\right)&=\left(\boldsymbol x^{\left(l\right)}\left(t\right)\right)^{\textrm{T}}{\boldsymbol U}^{\left(l\right)}\\
            &=\left(1-r^{2N}\right)r^{k}\left(\boldsymbol x^{\left(l\right)}\left(t\right)\right)^{\textrm{T}}\left({\boldsymbol A}^{\left(l\right)}\right)^{-1}{\textrm{rot}}_{k}\left(\boldsymbol V_{1\ldots N}^{\left(l\right)}\right)\label{parallel_yt}{\textrm{.}}
    \end{aligned}
  \end{displaymath}
  
  In the parallel architecture, the total output is defined as $y\left(t\right)=\sum_{l=1}^{L}y^{\left(l\right)}\left(t\right)/L$. Hence, the covariance of the output with the $k$-slot delayed source can be calculated as:
  \begin{equation}
    \begin{aligned}
        \textrm{Cov}\left(y\left(t\right),s\left(t-k\right)\right)=&\textrm{Cov}\left(\sum_{l=1}^{L}y^{\left(l\right)}\left(t\right)/L,s\left(t-k\right)\right)\\
                                        =&\sum_{l=1}^{L}\left(1-r^{2N}\right)r^{k}\\
                                        &\times \textrm{Cov}\left(x^{\left(l\right)}\left(t\right)^{\textrm{T}},s\left(t-k\right)\right)\\
                                        &\times \left({\boldsymbol A}^{\left(l\right)}\right)^{-1}{\textrm{rot}}_{k}\left(V_{1\ldots N}^{\left(l\right)}\right)/L\\
                                        =&\sum_{l=1}^{L}\left(1-r^{2N}\right)r^{2k}\sigma ^2\left({\textrm{rot}}_{k}\left(V_{1\ldots N}^{\left(l\right)}\right)\right)^{\textrm{T}}\\
                                        &\times \left({\boldsymbol A}^{\left(l\right)}\right)^{-1}{\textrm{rot}}_{k}\left(V_{1\ldots N}^{\left(l\right)}\right)/L\\
                                        =&\left(1-r^{2N}\right)r^{2k}\sigma ^2\zeta _{k}{\textrm{.}}
    \end{aligned}
  \end{equation}
  
  The variance of the observed output will be:
  \begin{equation}
    \begin{aligned}
        \textrm{Var}\left(y\left(t\right)\right)=&\textrm{Var}\left( \sum_{l=1}^{L}y^{\left(l\right)}\left(t\right)/L\right) \\
                                 =&\sum_{m=1}^{L}\sum_{n=1}^{L}{\mathbb E}\left( \left(y^{\left(m\right)}\left(t\right)\right)^{\textrm{T}}y^{\left(n\right)}\left(t\right)\right)/L^2{\textrm{,}}
    \end{aligned}
  \end{equation}
  where
  \begin{equation}
    \begin{aligned}
        {\mathbb E}&\left(\left(y^{\left(m\right)}\left(t\right)\right)^{\textrm{T}}y^{\left(n\right)}\left(t\right)\right)\\
                                 =&\left(\boldsymbol U^{\left(m\right)}\right)^{\textrm{T}}{\mathbb E}\left[\boldsymbol x^{\left(m\right)}\left(t\right)\left(\boldsymbol x^{\left(n\right)}\left(t\right)\right)^{\textrm{T}}\right]{\boldsymbol U^{\left(n\right)}}\\
                                 =&\frac{1-r^{2N}}{\sigma^2}\left(\boldsymbol p_k^{\left(m\right)}\right)^{\textrm{T}}\left(\Omega ^{\left(m\right)}\right)^{-1}\Gamma ^{-2}\left(\left(\Omega ^{\left(m\right)}\right)^{-1}\right)^{\textrm{T}}\\
                                 &\times\left(\Omega ^{\left(m\right)}\right)^{\textrm{T}}\Gamma ^2\Omega ^{\left(n\right)}\left(\Omega ^{\left(n\right)}\right)^{-1}\Gamma ^{-2}\left(\left(\Omega ^{\left(n\right)}\right)^{-1}\right)^{\textrm{T}}\boldsymbol p_k^{\left(n\right)}\\
                                =&\left(1-r^{2N}\right)\sigma ^2r^{2k}{\textrm{rot}_k}\left({\boldsymbol V^{\left(m\right)}_{1\ldots N}}\right)^{\textrm{T}}\left(\Omega ^{\left(m\right)}\right)^{-1}\Gamma ^{-2}\\
                                &\times\left(\left(\Omega ^{\left(n\right)}\right)^{-1}\right)^{\textrm{T}}{\textrm{rot}_k}\left({\boldsymbol V^{\left(n\right)}_{1\ldots N}}\right)\\
                                =&\left(1-r^{2N}\right)\sigma ^2r^{2k}{\textrm{rot}_k}\left({\boldsymbol e_1}\right)^{\textrm{T}}\Gamma ^{-2}{\textrm{rot}_k}\left({\boldsymbol e_1}\right)\\
                                =&\left(1-r^{2N}\right)\sigma ^2r^{2k}\zeta _k{\rm,}
    \end{aligned}
  \end{equation}
  and ${\mathbb E}\left[x^{\left(m\right)}\left(t\right)\left(x^{\left(n\right)}\left(t\right)\right)^{\textrm{T}}\right]$ has similar results to (\ref{parallel_x}).
  
  Hence,
  \begin{equation}
    \begin{aligned}
        \textrm{Var}\left(y\left(t\right)\right)&=\left(1-r^{2N}\right)r^{2k}\sigma ^2\zeta _{k}\\
            &=\textrm{Cov}\left(y\left(t\right),s\left(t-k\right)\right){\textrm{.}}
    \end{aligned}
  \end{equation}
  
  The squared correlation coefficient between the desired output $s(t-k)$ and the network output $y(t)$ is given by:
  \begin{equation}
    \begin{aligned}
        C_k&=\frac{\textrm{Cov}^2\left(y\left(t\right),s\left(t-k\right)\right)}{\textrm{Var}\left(s\left(t-k\right)\right)\textrm{Var}\left(y\left(t\right)\right)}\\
            &=\frac{\textrm{Var}\left(y\left(t\right)\right)}{\sigma ^2}\\
            &=\left(1-r^{2N}\right)r^{2k}\zeta _{k}.
    \end{aligned}
  \end{equation}
  
  The MC of the parallel ESN can be derived as follows:
  \begin{equation}
    \begin{aligned}
        C&=\sum_{k=1}^{\infty}C_k\\
          &=\sum_{k=0}^{\infty}C_{k}-C_0\\
          %&=\sum_{k=0}^{\infty}\left(1-r^{2N}\right)r^{2k}\zeta _{k}-\left(1-r^{2N}\right)r^0\zeta_0\\
          &=\left(1-r^{2N}\right)\sum_{k=0}^{N-1}r^{2k}\zeta _k\sum_{j=0}^{\infty}r^{2Nj}-\left(1-r^{2N}\right)\\
          &=\sum_{k=0}^{N-1}r^{2k}\zeta _k-\left(1-r^{2N}\right)\\
          &=N-1+r^{2N}.
    \end{aligned}
  \end{equation}
\end{IEEEproof}

From Theorem~\ref{theorem:parallel}, we can see that the theoretical MC of the parallel ESN increases with the reservoir size $N$ and reservoir weight $r$. However, the MC of parallel ESN is similar to that of ESN derived in \cite[Theorem 1]{Rodan2011Minimum}. From the perspective of the network structure, the parallel ESN duplicates $L$ ESNs and averages the results, instead of changing the internal structure of the ESN. The advantage of the parallel structure is that it can reduce the prediction error by averaging several reservoir outputs, thereby improving the prediction accuracy.

Theorem~\ref{theorem:parallel} also shows that the theoretical MC of an ESN can never exceed the reservoir size $N$, which is also the reservoir's maximum storage for recoding the input streams. The difference between the maximum storage and the theoretical MC implies that an ESN cannot record all the historical data from the input sequence. For the cascading architecture of the series ESN, $L$ reservoirs mean that the MC is constrained by each reservoir, i.e., it decreases $L$ times. In consequence, the MC of the series ESN will be smaller than that of the traditional shallow ESN and the parallel ESN. Due to space limitations, the theoretical analysis of the MC for the series ESN is left for future work.

\section{Simulation Results}\label{sec:exp}
\begin{table}[tbp]
\renewcommand{\arraystretch}{1.3}
\caption{Simulation Parameters}
\label{table:sim_para}
\centering
\begin{tabular}{c||c}
\hline
\textbf{Parameters} & \textbf{Values}\\
\hline\hline
$L$ & 3\\
\hline
$K$ & 1\\
\hline
$M$ & 1\\
\hline
$L_{\rm tr}$ & 500 (parallel)/700 (series)\\
\hline
$L_{\rm te}$ & 500 (parallel)/700 (series)\\
\hline
$L_{\rm fo}$ & 100\\
\hline
\end{tabular}
\end{table}
Next, we evaluate the prediction capability of the proposed deep ESNs by using the normalized root mean square error (NRMSE) metric:
\begin{equation}
    E=\sqrt \frac{\overline{\sum\left( \hat{\boldsymbol y}\left( t\right) -\boldsymbol y\left( t\right) \right)^2}}{\textrm{Var}\left(\boldsymbol y\left( t\right) \right)}\label{NRMSE}{\textrm{,}}
\end{equation}
where $\hat{\boldsymbol y}\left( t\right)$ is the predicted output, $\boldsymbol y\left( t\right)$ is the target output, $\overline \cdot$ is the mean operator, and $\textrm{Var}\left( \cdot\right)$ is the variance operator. 

The dataset is produced from the nonlinear autoregressive moving average (NARMA) system \cite{Atiya2000New}. The NARMA system is a discrete time system, whose current output depends on both the current input and the previous output. The nonlinearity and recursiveness of NARMA make it difficult to model. We use a fixed-order NARMA time series as the dataset:
\begin{equation}
    y\left(t\right)=0.7s\left( t-\tau\right)+\left (1-y\left( t-1\right) \right)y\left( t-1\right)+0.1\label{NARMA}{\textrm{,}}
\end{equation}
where $y\left( t\right)$ is the system output at time $t$, $s\left( t\right)$ is the system input at time $t$, that is generated randomly over a uniformly distribution in the range of $\left( 0,1\right)$, and $\tau$ captures the dependency length. We denote the length of the training and test sequences by $L_{\textrm{tr}}$ and $L_{\textrm{te}}$. The first $L_{\textrm{fo}}$ predicted outputs from the training and test sequences are ignored in the calculation of the NRMSE. This is because the reservoir is unstable during the initial training period. Hence, the predicted result during this period is meaningless. Here, we need to note that the amount of the ignored output in a series ESN should be $L$ times larger than that in a parallel ESN. This is because the data in a series ESN needs to enter $L$ reservoirs sequentially. Every time the data enters in a reservoir, the first $L_{\textrm{fo}}$ output will be inaccurate and should be ignored. Hence, the prediction accuracy of the series ESN can be comparable to the accuracy of the parallel ESN. Our detailed simulation parameters are listed in Table~\ref{table:sim_para}.

\begin{figure}[tbp]
    \centering
    \subfigure[Training set]{
    \label{figure:parallel_train}
    \includegraphics[width=2.4in]{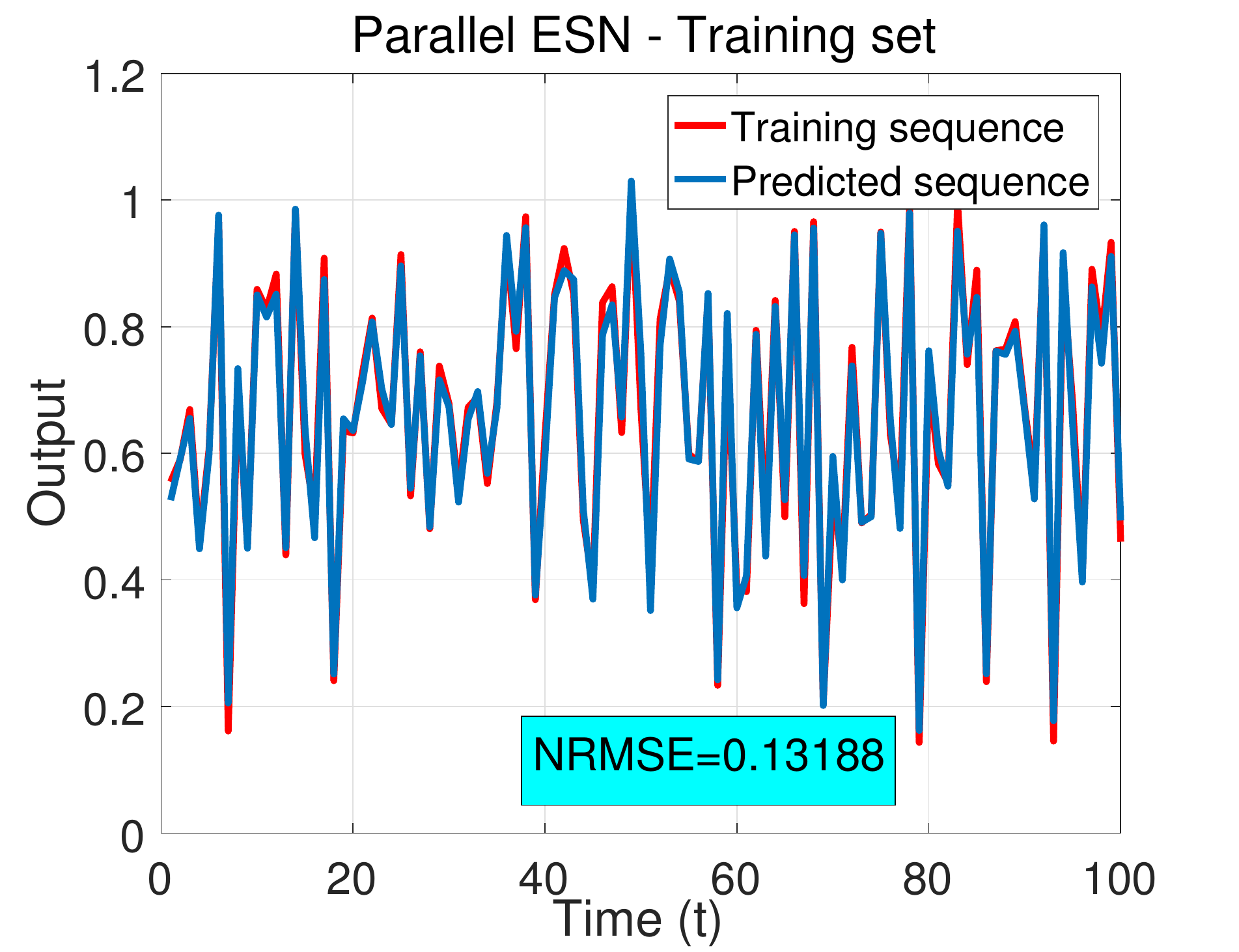}}
    \subfigure[Test set]{
    \label{figure:parallel_test}
    \includegraphics[width=2.4in]{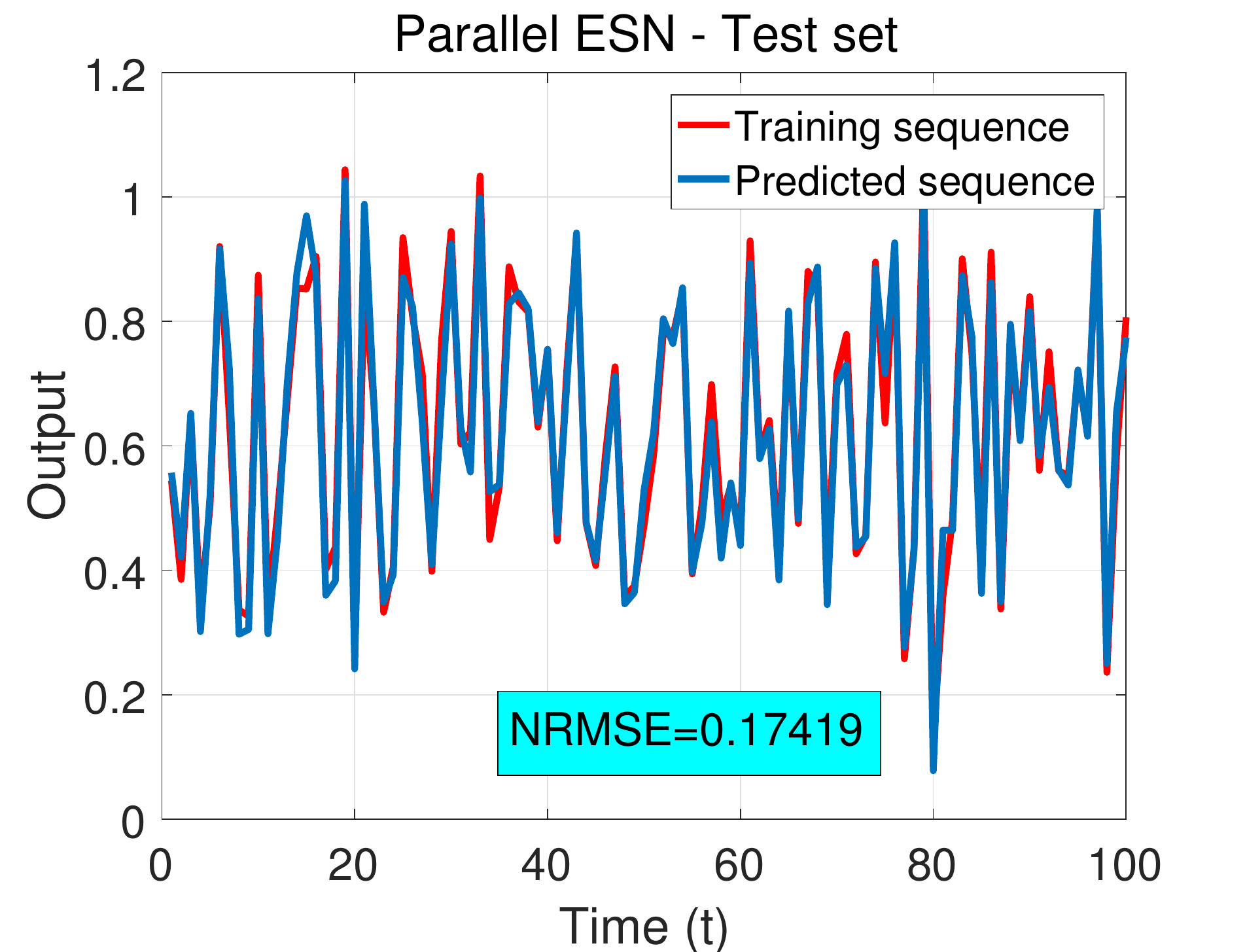}}
    \caption{Training result of a parallel ESN ($N=50$, $\tau =5$).}
    \label{figure:parallel_contrast}
\end{figure}
\begin{figure}[tbp]
    \centering
    \subfigure[Training set]{
    \label{figure:series_train}
    \includegraphics[width=2.4in]{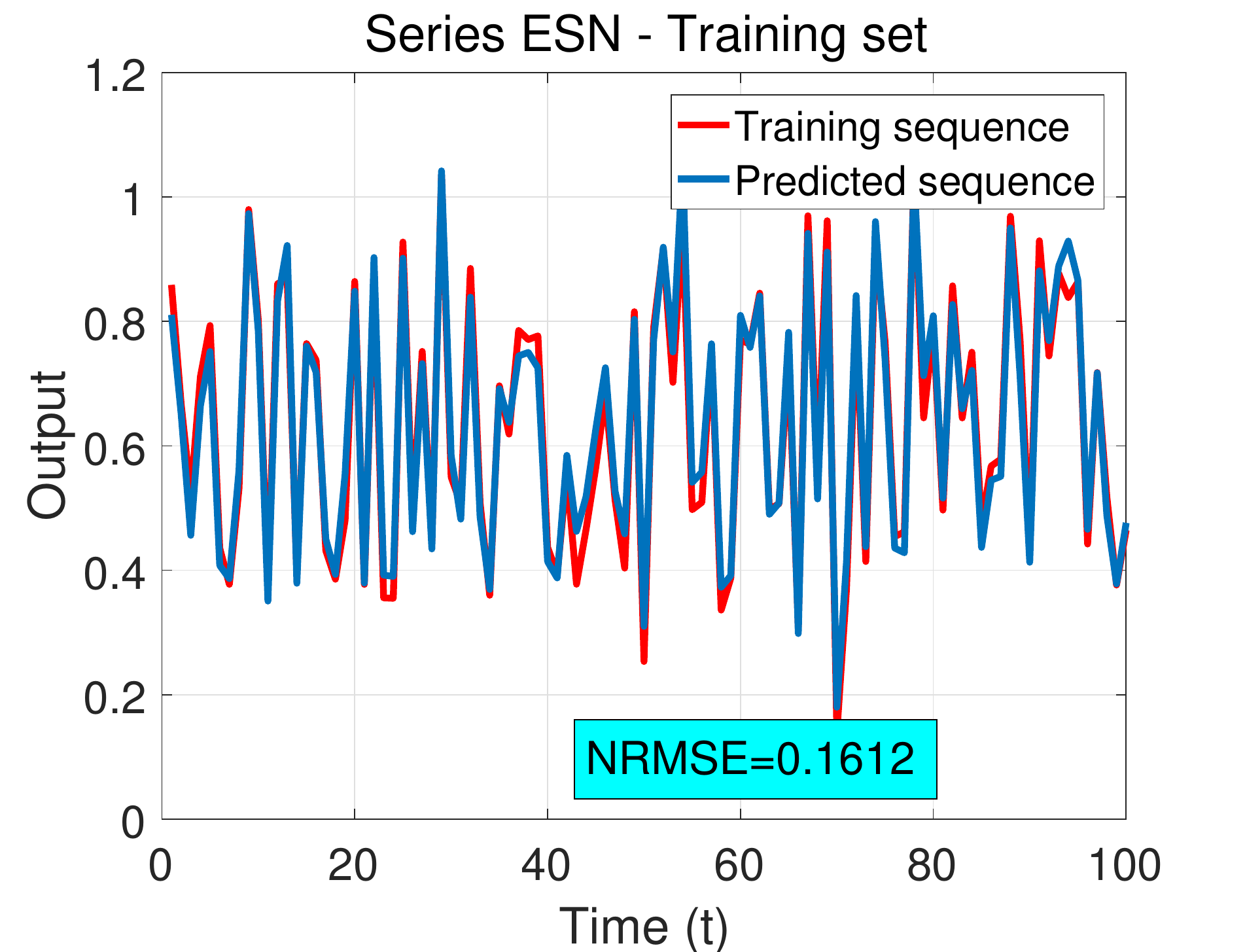}}
    \subfigure[Test set]{
    \label{figure:series_test}
    \includegraphics[width=2.4in]{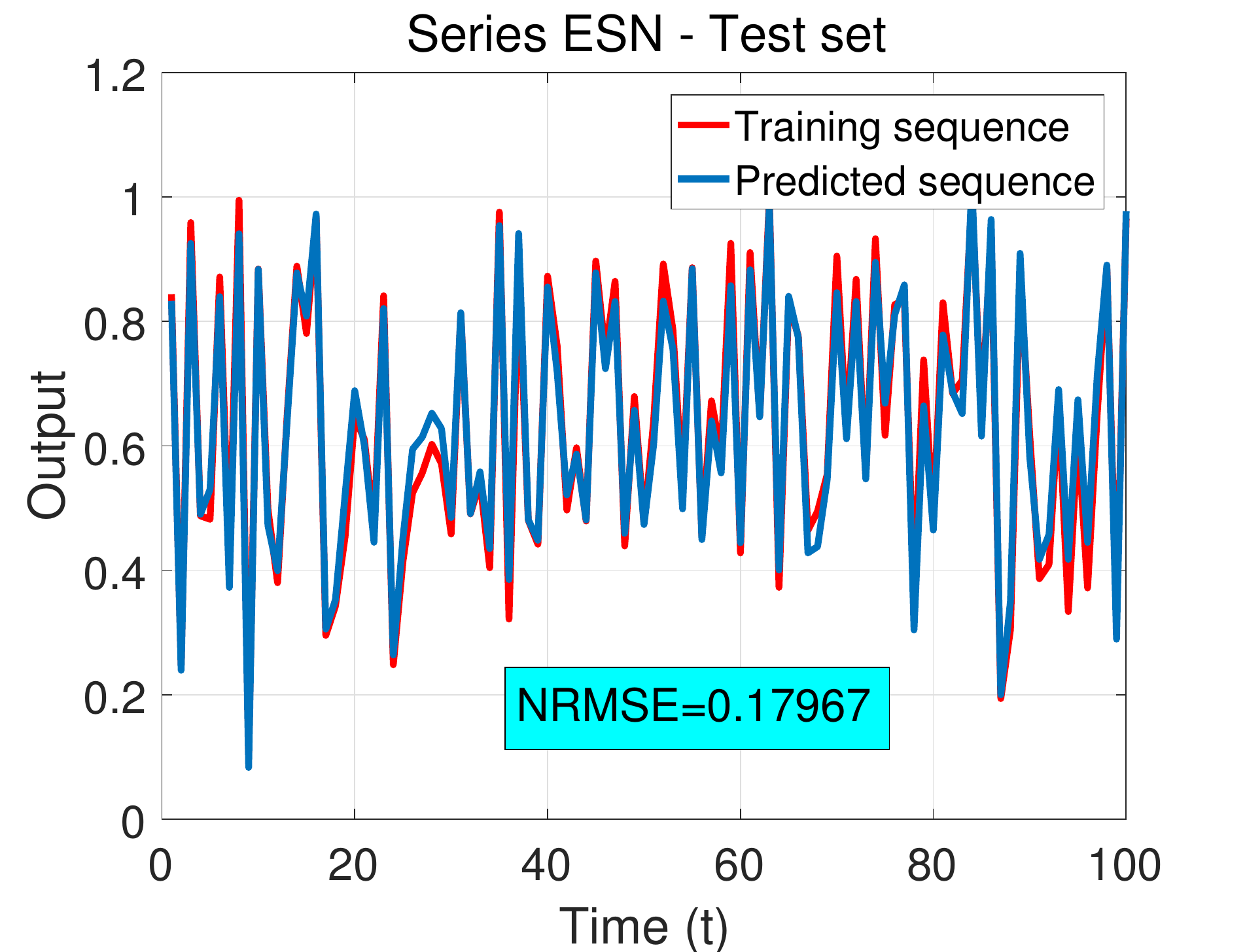}}
    \caption{Training result of a series ESN ($N=50$, $\tau =5$).}
    \label{figure:series_contrast}
\end{figure}

Figs.~\ref{figure:parallel_contrast} and \ref{figure:series_contrast} show the training results of a parallel ESN and a series ESN by contrasting the predicted sequence with the training sequence. Figs. \ref{figure:parallel_train} and \ref{figure:series_train} show that, when the training sets are tested, the predicted sequences almost fit the training sequences. This result shows that the parallel ESN and the series ESN are well-trained. In Figs.~\ref{figure:parallel_test} and \ref{figure:series_test}, we can see some gap between the predicted sequence and the training sequence. The NRMSE of the outputs also increases. This is due to the fact that, when using the test sets, the prediction accuracy is a little lower than when using the training sets. This reduction of the prediction accuracy reveals that the trained parallel ESN and the trained series ESN are not overfitted.

\begin{figure}[t]
\setlength{\belowcaptionskip}{-2em}
\centerline{\includegraphics[width=2.6in]{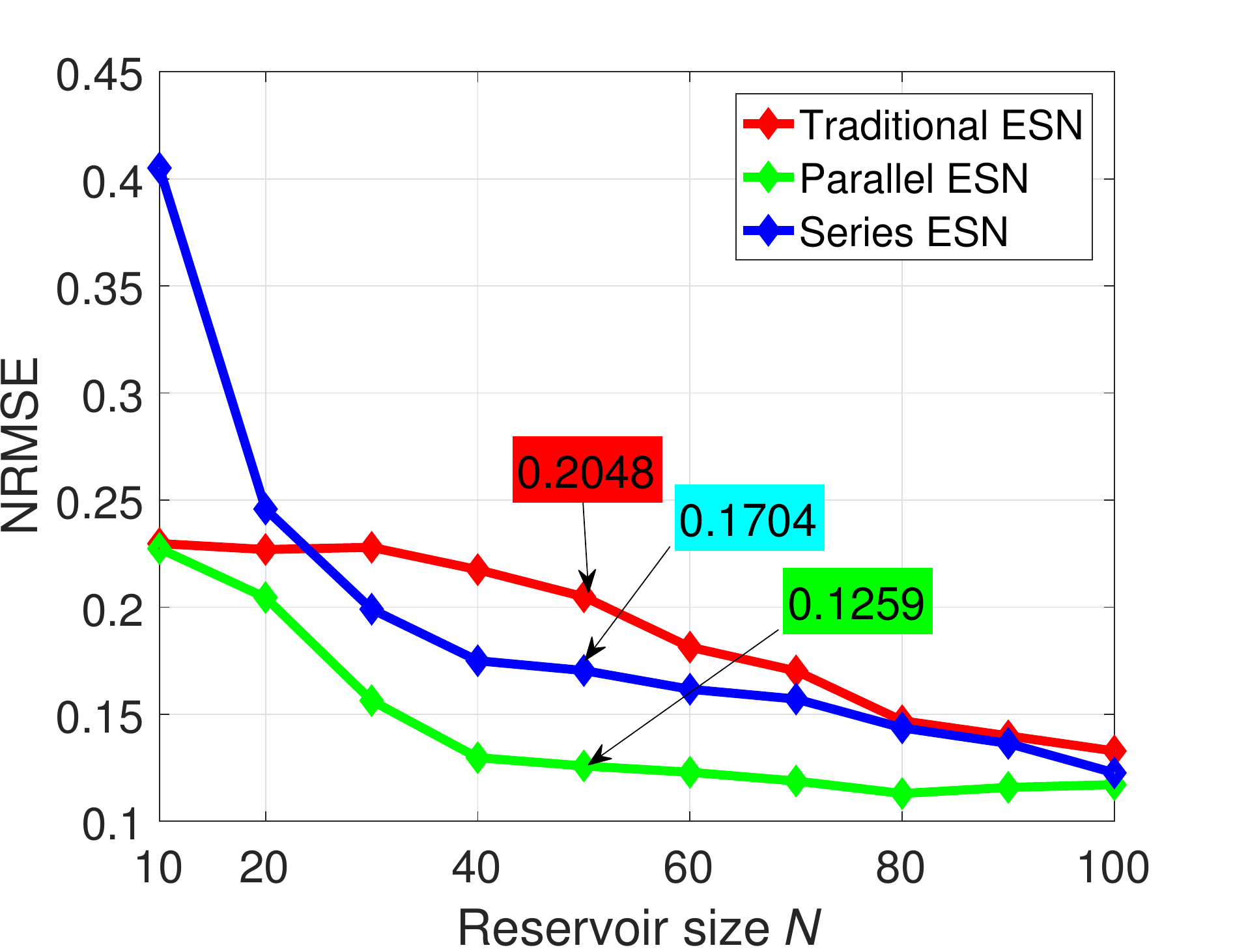}}
\caption{NRMSE as the reservoir size varies ($\tau=5$).}
\label{figure:neuron}
\end{figure}

\figurename~\ref{figure:neuron} shows that the NRMSE decreases as the reservoir size increases, for all considered ESN architectures. This is due to the fact that the ESN memory capacity is proportional to the reservoir size. \figurename~\ref{figure:neuron} also shows that parallel ESNs have the lowest NRMSE among the three kinds of ESNs. This is due to the fact that, compared to a traditional shallow ESN and a series ESN, a parallel ESN not only mitigates prediction error, but also has the best MC in all considered ESNs. From \figurename~\ref{figure:neuron}, we can see that the NRMSE of series ESNs becomes lower than that of single ESNs after the reservoir size increases to 30. This is due to the fact that the series architecture extracts new features and relations between input and output sequences so as to achieve a more accurate prediction. \figurename~\ref{figure:neuron} shows that the NRMSE of series ESNs is larger than the NRMSE of shallow ESNs when the reservoir size is 10 or 20. This is due to the fact that, when the reservoir size is small, the prediction accuracy of a single ESN is low so that the series ESN captures inaccurate features between the input and output sequences. These inaccurate features lead to a low prediction accuracy for the series ESN.
%The anomaly is due to that the series ESN not only captures more features than the shallow ESNs between the input and output sequences, it enhances the memory of prediction error produced by the first $L-1$ reservoirs as well. 
When the reservoir size is relatively small and the prediction result is relatively inaccurate, a prediction error will have a significant impact on the prediction performance. When the reservoir size is $N=50$, the NRMSE of the traditional ESN is 0.2048, while the NRMSE resulting from the parallel ESNs and the series ESNs will be 0.1259 and 0.1704. Compared to the traditional ESN, the parallel ESN achieves 38.5\% reduction and the series ESN achieves 16.8\% reduction in terms of the NRMSE.

\begin{figure}[tbp]
\vspace{-1.4em}
\setlength{\belowcaptionskip}{-2em}
\centerline{\includegraphics[width=2.6in]{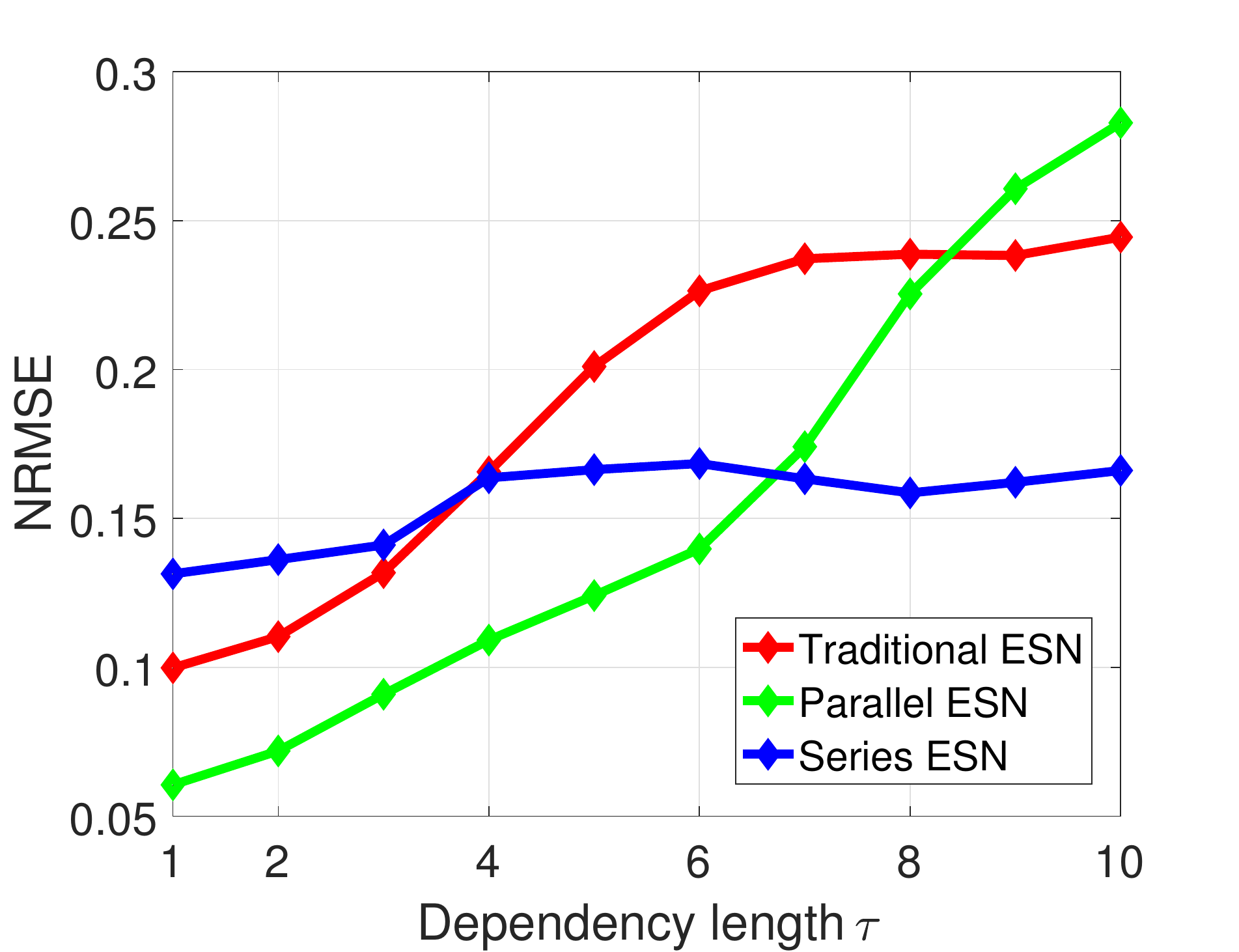}}
\caption{NRMSE as the dependency length $\tau$ varies ($N=50$).}
\label{figure:tau}
\end{figure}

\figurename~\ref{figure:tau} shows that the prediction accuracy of the all considered ESNs changes as the dependency length $\tau$ of the NARMA system varies.
%$\tau$ indicates the order of the NARMA system, which is proportional to the system complexity. For a certain ESN, the internal structure is fixed and the MC is determined by the structure. 
In \figurename~\ref{figure:tau}, as the dependency length $\tau$ increases, the prediction accuracy decreases. This is due to the fact that a larger dependency length $\tau$ leads to a more complex NARMA system with a higher order. With the same internal structure, the prediction accuracy of an ESN decreases as the target system becomes more complex. \figurename~\ref{figure:tau} also shows that series ESNs has the lowest NRMSE as $\tau$ increases. This is due to the fact that the series ESN cascades multiple reservoirs, so that more features and relations between the input and the output are extracted. Based on that, the series ESN yields higher prediction accuracy for more complex systems compared to the traditional ESN and the parallel ESN.
\begin{figure}[t]
\setlength{\belowcaptionskip}{-2em}
\centerline{\includegraphics[width=2.6in]{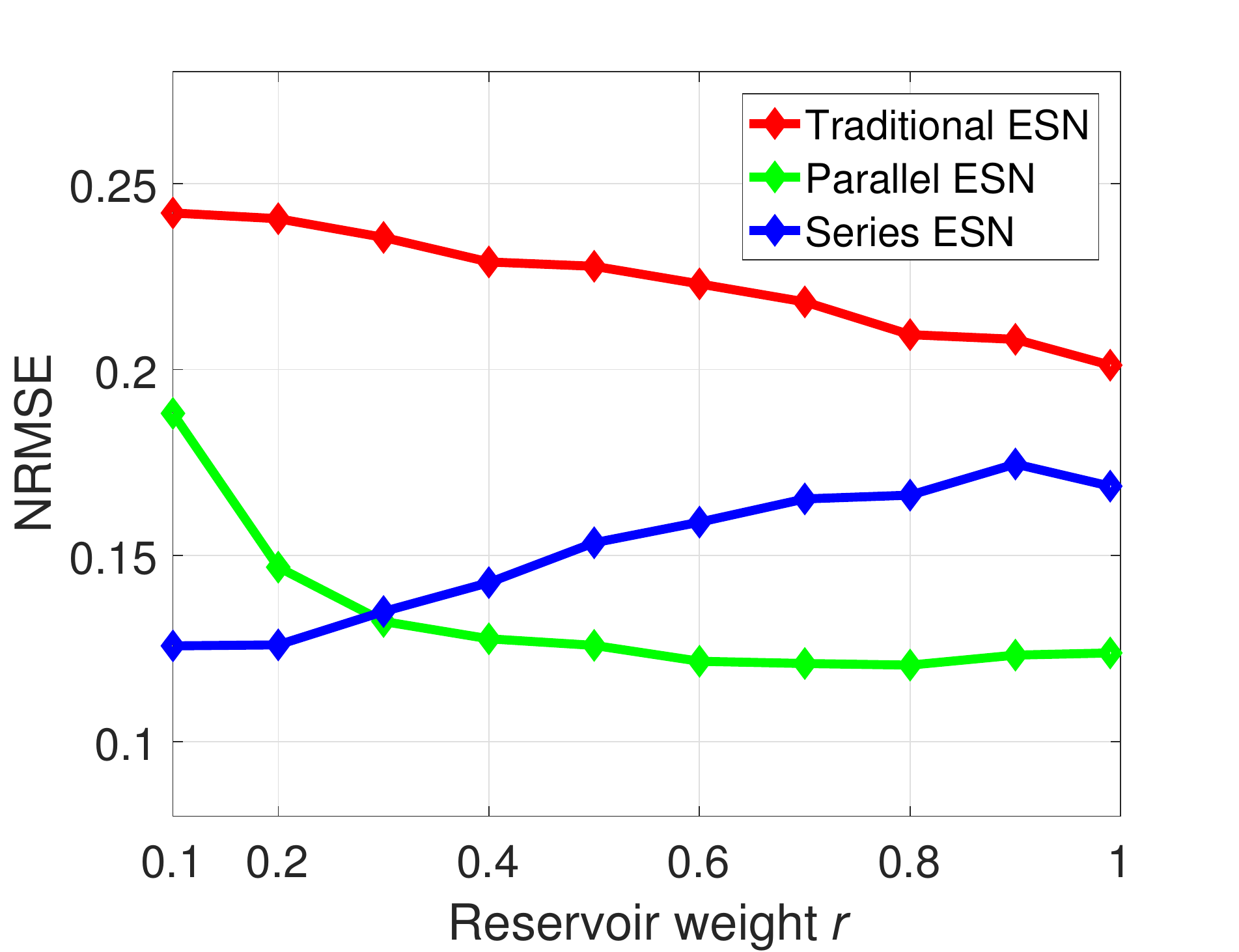}}
\caption{NRMSE as the reservoir weight $r$ varies ($N=50$).}
\label{figure:r}
\end{figure}
\figurename~\ref{figure:r} shows that the NRMSE of traditional ESNs and parallel ESNs decrease as the reservoir weight $r$ increases, but the NRMSE of series ESNs increases. This is because the MC of parallel ESNs is proportional to $r^{2N}$ as well as the MC of traditional ESNs.
%However, the MC of series ESNs has a penalty term related to $r$. Note that the number of reservoirs is 3 in this experiment. The increasing of $r$ leads to the decline of the series ESN MC. 
%We can conclude that the MC, which is based on the architecture of the network, determines the prediction accuracy of the network to a certain extent.

\section{Conclusion}\label{sec:con}
In this paper, we have proposed two novel deep ESN architectures, parallel ESN and series ESN. Compared to a traditional shallow ESN, a parallel ESN decreases the prediction error by averaging multiple separate reservoir outputs, while the series ESN captures new features to predict the system output by cascaded training. We have also analyzed the MC of the parallel deep ESN and we have shown that the MC does not exceed and is arbitrarily close to the reservoir size.
Simulation results show that deep ESNs can decrease the prediction error compared to a traditional shallow ESN. In particular, the parallel ESN yields a 38.5\% reduction in terms of the NRMSE, and the series ESN yields a 16.8\% reduction. 

\def\baselinestretch{0.70}

\end{document}